\documentclass[10pt,a4paper]{article}

\usepackage[utf8]{inputenc}   
\usepackage[T1]{fontenc}          
\usepackage{graphicx}                   
\usepackage{float}                      
\usepackage{geometry}                   
\usepackage{hyperref}                   
\usepackage{setspace}                   
\usepackage{color}                      
\usepackage[table]{xcolor}              
\usepackage{fancyhdr}                   
\usepackage{fourier-orns}               
\usepackage[usenames,dvipsnames]{pstricks}
\usepackage{listings}

\hypersetup{                            
        colorlinks=true,                
        breaklinks=true,                
        urlcolor=black,                 
        linkcolor=black,                
        pdftitle={},                    
        pdfauthor={O. Bailleux},        
        pdfproducer={O. Bailleux},      
        pdfsubject={PPC},               
        pdftoolbar=true,                
        pdfmenubar=true,                
        pdfstartview={FitH},            
}

\sffamily                               
\title{\vspace{-1cm}\textsc{BoolVar/pb} \texttt{v1.0}, a java library for translating\\ pseudo-Boolean constraints into \textsc{cnf} formulae}

\normalsize\author{Olivier Bailleux, Université de Bourgogne}

\usepackage{amssymb}

\begin{document}

\sloppy 

\rfoot{\small page \thepage / \pageref{pageend}}

\maketitle 




\begin{abstract}
\textsc{BoolVar/pb} \texttt{v1.0} is an open source \texttt{java} library dedicated to the translation of pseudo-Boolean constraints into \textsc{cnf} formulae. Input constraints can be categorized with tags. Several encoding schemes are implemented in a way that each input constraint can be translated using one or several encoders, according to the related tags. The library can be easily extended by adding new encoders and / or new output formats. It is available at \url{http://boolvar.sourceforge.net/}.
\end{abstract}

\section{Introduction}

Let us begin by an introductory exemple, which consists in translating the constraint $5 x_1 + 3 \overline{x_2} + x_3 \leq 8$ into a \textsc{cnf} formula. At first, an input model must be created as follows:

\begin{lstlisting}
    InputModel m = new InputModel();
\end{lstlisting}

Then, the three variables $x_1, x_2, x_3$ can be created as an array of instances of the class \texttt{Variable}.

\begin{lstlisting}
    Variable[] x = new Variable[3];
    for(int i=0; i<3; i++) x[i] =  new Variable();
\end{lstlisting}

The input constraint is composed of literals that can be produced from the variables. Each literal is an instance of the class \texttt{Literal}.

\begin{lstlisting}
    Literal[] lits = new Literal[3];
    lits[0] = x[0].getPosLit();
    lits[1] = x[1].getNegLit();
    lits[2] = x[2].getPosLit();
\end{lstlisting}

Before to create the input constraint, we have to define the coefficients related to each literal and the tag that will be assigned to this constraint. Note that it is allowed to assign several tags to the same constraint. For example, \texttt{setTags(1,3)} will assigns the tags 1 and 3 to the upcoming constraints that will be created until the next call of \texttt{setTags}.

\begin{lstlisting}
   int[] coeffs = {5,3,1};
   setTags(1);
\end{lstlisting}

To complete the building of the input model, it only remains to create the input constraint, which is simplified by using the static factory method \texttt{makeLeq}, and add it to the model.

\begin{lstlisting}
    Constraint q = makeLeq(coeffs,lits,8);
    m.addConstraint(q);
\end{lstlisting}

At this time, the input model is created. It can be print to the screen for verification purpose.

\begin{lstlisting}
    System.out.println(m.toString());
\end{lstlisting}

Now, the input model must be translated to an output problem, namely a \textsc{cnf} formula. This suppose to create an instance of the class \texttt{CNFProlem} in the following way:

\begin{lstlisting}
    OutputProblem out = new CNFProblem();
\end{lstlisting}

Now, suppose that we want to encode the input constraint (which is marked with the tag 1) redundantly with two encoders. We have to create an instance of each of these encoders, and to assign these encoders to the tag 1. As a consequence, each constraint of the input model tagged with 1 will be translated using these two encoders.

\begin{lstlisting}
    Encoder2cnf direct = new CNFdirectEncoder();
    Encoder2cnf bdd = new CNFbddEncoder();
    out.assignEncoder(1,direct);
    out.assignEncoder(1,bdd);
\end{lstlisting}

The translation process will be achieved by reading the input model thanks to the method \texttt{read} of the output problem.

\begin{lstlisting}
    out.read(m);
\end{lstlisting}

The resulting \textsc{cnf} formula is given in \textsc{dimacs} format by the method \texttt{getOutput}.

\begin{lstlisting}
    System.out.print(out.getOutput());
\end{lstlisting}

For example, if only the encoder \texttt{CNFdirectEncoder} is used, the result is the following:

\begin{lstlisting}
    p cnf 3 1
    -1 2 -3 0
\end{lstlisting}

This encoder produces an exponential number of clauses in the general case, but can produce quite compact outputs from small input constraints. The currently available encoders will be described in section \ref{sect-encoders}.

To conclude this brief presentation, let us mention that \textsc{BoolVar} also provide a class \texttt{PBproblem} and an encoder \texttt{PBbasicEncoder} allowing to produce the output problem as an instance of pseudo-Boolean satisfiability with the \textsc{opb} format. This allows a same input problem to be solved either using a \textsc{sat} solver and a pseudo-Boolean satisfiability solver, in order to compare the performances and the relevance of the two approaches.

\section{Description}

This section presents the main aspects and resources of \textsc{BoolVar/pb} from the user side and as well as a short description of its internal structure.

\subsection{Architecture}

\begin{figure}[position]
\begin{center}
\includegraphics[scale=0.4]{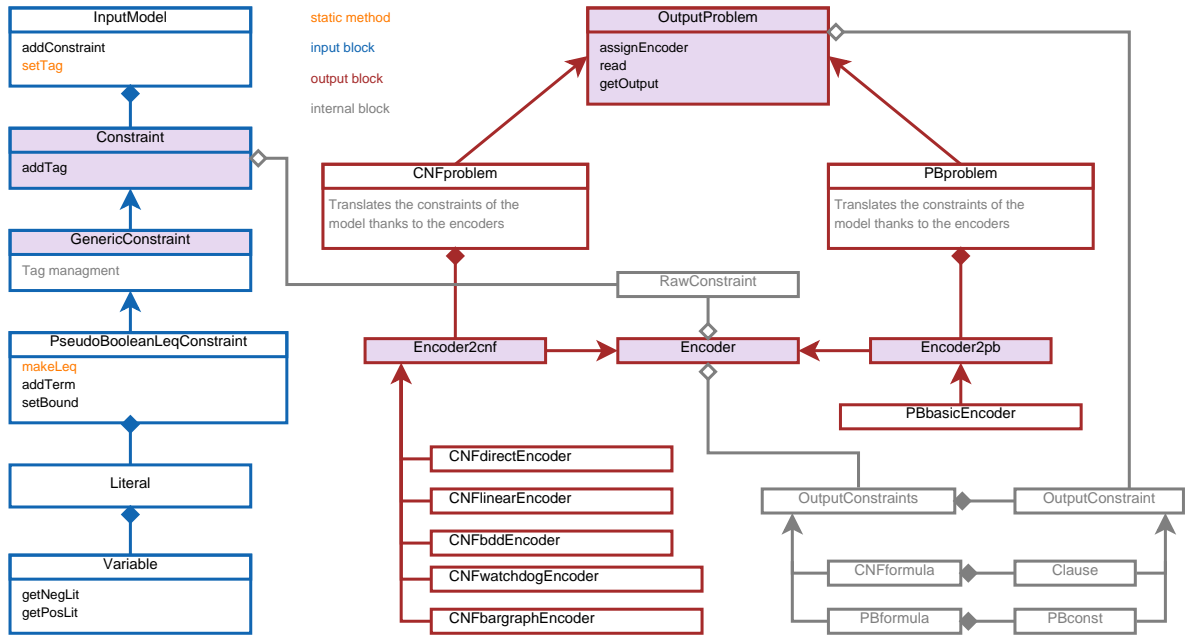}
\end{center}
\caption{The architecture of \textsc{BoolVar/pb}\label{fig-archi}}
\end{figure}

Figure \ref{fig-archi} presents the architecture of the \textsc{BoolVar/pb} library, which can be decomposed in three parts: the input block, the output block, ans the internal block.

\subsubsection{The input block}

This part includes the classes allowing the user to create and specify an input problem.

\begin{description}
\item[\texttt{InputModel}]~
 
The container for the input constraints. Its constructor must be used to create a new input problem where the input constraints will be added thank to the method \texttt{addConstraint}.

\item[\texttt{Constraint}]~

An interface requiring that any input constraint implements a method \texttt{addTag}, which allows to assign tags to constraints.

\item[\texttt{GenericConstraint}]~

An abstract class which implements the tag management system, which is the same for any input constraint.

\item[\texttt{PseudoBooleanLeqConstraint}]~

A pseudo-Boolean inequality constraint of the form $a_1 x_1 + \cdots + a_n x_n \leq b$, where $x_1, \ldots, x_n$ are propositional literals (i.e., Boolean variables or negated Boolean variables), $a_1, \ldots, a_n$ and $b_n$ are positive integers. The instances of this class can be created in two ways : (1) using the provided constructors and building methods, or (2) using the static factory method \texttt{Boolvar.makeLeq}.

\item[\texttt{Literal}]~

The building block for input constraints. A literal can be produces either by using the constructor of this class, or by using the methods \texttt{getNegLit} and \texttt{getPosLit} of the class \texttt{Variable}.

\item[\texttt{Variable}]~

A representation for the propositional variables that are used both in the input and in the output constraints.
\end{description}

\subsubsection{The output block}

This part includes the classes allowing the user to specify the output problem, as wall as the way to produce this problem from the input constraints.

\begin{description}
\item[\texttt{OutputProblem}]~
 
This interface specify the methods that must be implemented in any kind of output problems: \texttt{assignEncoder}, \texttt{read}, and \texttt{getOutput}.

\item[\texttt{CNFproblem}]~

A representation for \textsc{cnf} output problems. Basically, any instance of this class contains a set of encoders and a string that will be receive the result of the translation. The internal methods ensures the translation process.

\item[\texttt{PBproblem}]~

A representation of a pseudo-Boolean output problem in the same way of the class \texttt{CNFproblem}. This class allows to produce an instance of the input problem in a form that allows to solve it using a pseudo-Boolean satisfiability solver.

\item[\texttt{Encoder}]~

This interface represents an encoder, i.e., a class that contains the resources for translating constraints.

\item[\texttt{Encoder2cnf}]~

This interface represents an encoder which produces a \textsc{cnf} formula as output. The available implementations will be presented in section \ref{sect-encoders}.

\item[\texttt{Encoder2pb}]~

This interface represents an encoder which produces a list of pseudo-Boolean constraints as output.

\item[\texttt{PBbasicEncoder}]~

This encoder puts the input problem in a format that allows it to be solved using a pseudo-Boolean satisfiability solver.
\end{description}

\subsubsection{The internal block}

This part includes the internal resources allowing the communication between the input and the output block, as well as the implementation of the translation process.

\begin{description}
\item[\texttt{RawConstraint}]~
 
The internal representation for input constraints. Any class which implements the interface \texttt{Constraint} must provide a method producing an array of such internal constraints. Currently, any instance of the class\texttt{ PseudoBooleanLeqConstraint} produces only one instance of \texttt{RawConstraint}, but a class \texttt{PseudoBooleanEqConstraint} could be implemented in a way to manage input constraints of the form $a_1 x_1 + \cdots + a_n x_n = b$. Such a constraint could either be represented as two internal inequality constraint or one internal equality constraint.

\item[\texttt{OutputConstraint}]~

A generic output constraint.

\item[\texttt{Clause}]~

Implementation of \texttt{OutpuConstraint} as a propositional clause. Contains resources that can be used by the \textsc{cnf} encoders to produce output clauses, which will be converted into strings by the class \texttt{CNFproblem}.

\item[\texttt{PBconst}]~

Implementation of \texttt{OutpuConstraint} as a pseudo-Boolean constraint. Contains resources that can be used by the \textsc{pb} encoder to produce output pseudo-Boolean constraints, which will be converted into strings by the class \texttt{CNFproblem}.

\item[\texttt{OutputConstraints}]~

A set of output constraints resulting from the translation process, which are aimed to be converted into strings to produce the output problem.

\item[\texttt{CNFformula}]~

A set of Clauses represented as instances of the class \texttt{Clause}.

\item[\texttt{PBformula}]~

A set of pseudo-Boolean constraints represented as instances of the class \texttt{PBconst}.
\end{description}

\subsection{User resources}

This section presents the main classes and methods allowing one to implement the resources provided by \textsc{BoolVar/pb}.

\subsubsection{Input block user resources}

\begin{description}
\item[\texttt{InputModel.InputModel()}]~
 
Create a new input model, which can be seen as a container for the input constraints. 

\item[\texttt{Variable.Variable()}]~
 
Create a new propositional variable, which is aimed to be used in an input constraint.

\item[\texttt{Variable.getPosLit()}]~
 
Returns a positive literal $v$ from the current variable $v$. If the literal $v$ does not exists, it is created thanks to the constructor of the class \texttt{Literal}, else the reference of the existing literal is returned. $v$ is aimed to be used in input constraints.

\item[\texttt{Variable.getNegLit()}]~
 
Returns a negative $\overline{v}$ literal from the current variable $v$. If $\overline{v}$ does not exists, it is created thanks to the constructor of the class \texttt{Literal}, else the reference of the existing literal is returned. $\overline{v}$ is aimed to be used in input constraints.

\item[\texttt{Variable.getLit(boolean sign)}]~
 
Returns a literal $v$ or $\overline{v}$, according to the value of \texttt{sign}, from the current variable $v$. If the required literal does not exists, it is created thanks to the constructor of the class \texttt{Literal}, else the reference of the existing literal is returned. 

\item[\texttt{Literal.Literal(Variable v, Boolean s)}]~
 
Create a new literal, which is aimed to be used in an input constraint. In order to avoid redundancies, literals must be preferably created thanks to the methods \texttt{getNegLit} and \texttt{getPosLit} provided by the class \texttt{Variable}.

\item[\texttt{BoolVar.setTag(int tag1)}]
\item[\texttt{BoolVar.setTag(int tag1,int tag2)}]
\item[\texttt{BoolVar.setTag(int tag1,int tag2,...,tag4)}]~

Sets the tags that will be assigned to the input constraints that will be created before the next call of \texttt{setTag}. The tags are arbitrary integers that will be assigned to encoders in a way to specify which encoder(s) must be used for translating each input constraint.

\item[\texttt{BoolVar.makeLeq(int[] c, Literal[] l, int b)}]
\item[\texttt{BoolVar.makeLeq(BigInteger[] c, Literal[] l, BigInteger b)}]~ 

Create a new pseudo-Boolean inequality constraint $\mathtt{c}[0] \mathtt{l}[0] + \cdots + \mathtt{c}[n-1] \mathtt{l}[n-1] \leq b$ (where $n$ is the size of the arrays \texttt{c} and \texttt{l}) from an array \texttt{c} of coefficients, an array \texttt{l} of literals, and a bound \texttt{b}. The resulting constraint is aimed to be added to the input model tanks to the method \texttt{InputModel.addConstraint}.

\item[\texttt{InputModel.addConstraint(Constraint q)}]~
 
Adds a new constraint to the current input model.
\end{description}

\subsubsection{Output block user resources}

\begin{description}
\item[\texttt{PBproblem.PBproblem()}]~
 
Creates a new output problem as a pseudo-Boolean satisfiability problem.

\item[\texttt{PBproblem.PBproblem()}]~
 
Creates a new output problem as a propositional satisfiability problem.

\item[\texttt{PBproblem.assignEncoder(int tag, Encoder x)}]~
 
Assigns the encoder \texttt{x} to the given tag. Several encoders can be assigned to the same tag by multiple calls to this method.

\item[\texttt{PBproblem.read(InputModel m)}]~
 
Reads the input model \texttt{m} in a way to translate each input constraint of \texttt{m} with the related encoders. This method must be used only one time, after all the input constraints are added to the input model.

\item[\texttt{PBproblem.getOutput()}]~
 
Returns the output problem as a string. This method must be called after the method \texttt{InputModel.read}.
\end{description}

\section{Available encodings \label{sect-encoders}}

The current version of \textsc{BoolVar/pb} includes 5 \textsc{cnf} encoders for pseudo-Boolean inequality constraints.

The underlying encoding methods can be classified in different categories with respect to the size of the output formula and the inference power of unit propagation (which is the basic filtering technique used in the \textsc{sat} solvers) on this formula.

Any \textsc{cnf} encoding which produces a \textsc{cnf} formula polynomially sized (exponentially sized, respectively) with respect to the number of variables in the input constraint is said to be \emph{polynomial} (\emph{exponential}, respectively).

Any \textsc{cnf} encoding is said to be a \textsc{pac} (like propagating arc consistency) encoding if and only if applying unit propagation on the resulting formula fixes the same variables as restoring arc consistency on the corresponding input constraint. It is said to be a \textsc{pic} (like propagating inconsistency) encoding if and only if applying unit propagation on the resulting formula produces the empty clause if   restoring arc consistency on the corresponding input constraint detects an inconsistency. Any \textsc{pac} encoding is necessarily a \textsc{pic} one.

\subsection{\texttt{CNFdirectEncoder}}

This simple exponential and \textsc{pac} encoding is briefly described in \cite{ DBLP:conf/sat/BailleuxBR09}. It can be seen as a variant of the \textsc{bdd} based encoding introduced in \cite{DBLP:journals/jsat/BailleuxBR06} where each path of the \textsc{bdd} is encoded with a clause in a way that no additional variable is required.

\subsection{\texttt{CNFbddEncoder}}

This exponential \textsc{pac} \textsc{bdd}-based encoding was introduced in \cite{DBLP:journals/jsat/BailleuxBR06}. It generally produces a smaller formula that the direct encoding, because it uses additional variables corresponding to each node of the \textsc{bdd}, which allows to factorize identical subgraphs. Unlike the direct one, this encoding is polynomial for cardinality constraints, i.e., when all the coefficients are 1.

\subsection{\texttt{CNFlinearEncoder}}

This linear encoding is described in \cite{DBLP:journals/jsat/EenS06}. It is inspired from the encoding introduced in \cite{warner68}. It is neither \textsc{pac} nor \textsc{pic}. His main interest is the size of the produced formulae.

\subsection{\texttt{CNFwatchdogEncoder}}

This encoding, introduced in \cite{DBLP:conf/sat/BailleuxBR09}, is both polynomial and \textsc{pac} (then \textsc{pic}), but can sometimes produce output formulae of prohibitive size.

\subsection{\texttt{CNFbargraphEncoder}}

This is a variant of the watchdog encoding, also presented in \cite{ DBLP:conf/sat/BailleuxBR09}, which is \textsc{pic} but not \textsc{pac}, and produces smallest formulae.

\section{A commented example}

As an exemple, we will encode the bin-packing problem, where $n$ objects, each of them with a weight $w_j, 0 \leq j \leq n-1$, must be put into $m$ boxes with capacities $c_i, 0 \leq i \leq m-1$, in such a way that each object occurs in exactly one box and the sum of the weights of all the objects belonging to any box does not exceed the capacity of this box.

Each instance of this problem will be represented thanks to a matrix $v$ of Boolean variables, where $v_{i,j}=1$ means that the object $j$ is in the box $i$.

There are two kinds of constraints, namely:

\begin{itemize}
\item 
the unicity constraints, which ensures that each object belong to exactly one boxe:
\[
\forall j \in 0..n-1, \sum_{i=0}^{m}{v_{i,j}} = 1
\]

\item
the capacity constraints, ensuring that the sum of the weights of the objects in any box does not exceed the capacity of this box:
\[
\forall i \in 0..m-1, \sum_{j=0}^{n}{w_j v_{i,j}} \leq c_i
\]
\end{itemize}

The bin-packing problem can be encoded with \textsc{BoolVar} in the following way.
The integer arrays \texttt{weights} and \texttt{capacities} are supposed to contain the weights of the objects and the capacities of the boxes, respectively.

\begin{enumerate}
\item
Create the input model.

\begin{lstlisting}
InputModel p = new InputModel();
\end{lstlisting}

\item
Create and initialize the matrix of domain variables. 
\begin{lstlisting}
Variable[][] v = new Variable[m][n];
for(int i=0; i<m; i++)
    for(int j=0; j<n; j++)
        v[i][j] = new Variable();
\end{lstlisting}

\item
Create the unicity constraints and assign them the tag 1 with the method \texttt{setTag}. Each equality constraint $v_{0,j} + \cdots + v_{m,j} = 1$ is encoded as two inequality constraints $v_{0,j} + \cdots + v_{m,j} \leq 1$ and $\overline{v_{0,j}} + \cdots + \overline{v_{m,j}} \leq m-1$.

\begin{lstlisting}
setTags(1);
int[] coeffs = new int[m];
for(int i=0; i<m; i++) coeffs[i]=1;
for(int j=0; j<n; j++)
{
   Literal[] poslits = new Literal[m];
   Literal[] neglits = new Literal[m];
   for(int i=0; i<m; i++)
   {
       poslits[i] = v[i][j].getPosLit();
       neglits[i] = v[i][j].getNegLit();
   }
   p.addConstraint(makeLeq(coeffs,poslits,1));
   p.addConstraint(makeLeq(coeffs,neglits,m-1));
}
\end{lstlisting}

\item
Create the capacity constraints and assign them the tag 2.

\begin{lstlisting}
setTags(2);
for(int i=0; i<m; i++)
{
   Literal[] lits = new Literal[n];
   for(int j=0; j<n; j++)
       lits[j] = v[i][j].getPosLit();
   p.addConstraint(makeLeq(weights,lits,capacities[i]));
}
\end{lstlisting}

\item
Create the output problem.

\begin{lstlisting}
OutputProblem out = new CNFProblem();
\end{lstlisting}

\item
Create the two encoders that will be used to translate the input constraints: the \textsc{bdd} encoder will be used to translate the unicity constraints (tagged with 1), and the bargraph encoder will be used to translate the capacity constraints (tagged with 2).

\begin{lstlisting}
Encoder2cnf bdd = new CNFbddEncoder();
Encoder2cnf bg = new CNFbargraphEncoder();
out.assignEncoder(1,bdd);
out.assignEncoder(2,bg);
\end{lstlisting}

\item
Read the input model, which runs the translation process, and print the result as a string.

\begin{lstlisting}
out.read(p);
System.out.println(out.getOutput());
\end{lstlisting}
\end{enumerate}

\section{Perspectives}

The following evolutions are planned.

\subsubsection*{Predicting the size of the input formulae}

The size of the formula resulting from the translation of each input constraint is a critical parameter for the choice of the encodings. The interface \texttt{Encoder} will include a method providing this information.

\subsubsection*{Automatic choice of the encoders}

Assigning a dedicated tag to any constraint will ensure that the encoder to use for translating this constraint will be automatically selected accordingly to both the size of the output formula and the inference power of unit propagation on this formula.

\subsubsection*{Clauses as input constraints}

It will be allowed to add clauses as input constraints. These clauses will be directly added to the output problem, in a way to allow to deal with input problems that are specified with both clauses and pseudo-Boolean formulae.

\subsubsection*{Supporting new encodings}

Some encodings dedicated to cardinality constraints will be added. Because cardinality constraints are a special case of pseudo-Boolean ones, the pseudo-Boolean encoders can of course deal with cardinality constraints. But there exists specific encodings which could be more efficient and/or compact for cardinality constraints.

In addition, some of the already implemented encodings could be improved and / or hybridized in a way to reduce the size / efficiency ratio of the resulting output formulae.

\subsubsection*{Coupling with a solver}

The goal is to provide the resources for solving the output problem thanks to the solver sat4J \cite{sat4j}, in a way to build stand alone applications.

\label{pageend}



\bibliographystyle{plain}
\bibliography{boolvar}

\begin{thebibliography}{1}

\bibitem{DBLP:journals/jsat/BailleuxBR06}
Olivier Bailleux, Yacine Boufkhad, and Olivier Roussel.
\newblock A translation of pseudo boolean constraints to sat.
\newblock {\em JSAT}, 2(1-4):191--200, 2006.

\bibitem{DBLP:conf/sat/BailleuxBR09}
Olivier Bailleux, Yacine Boufkhad, and Olivier Roussel.
\newblock New encodings of pseudo-boolean constraints into cnf.
\newblock In {\em SAT}, pages 181--194, 2009.

\bibitem{sat4j}
Daniel~Le Berre and Anne Parrain.
\newblock The sat4j library, release 2.2.
\newblock {\em JSAT}, 7:59--64, 2010.

\bibitem{DBLP:journals/jsat/EenS06}
Niklas E{\'e}n and Niklas S{\"o}rensson.
\newblock Translating pseudo-boolean constraints into sat.
\newblock {\em JSAT}, 2(1-4):1--26, 2006.

\bibitem{warner68}
J.~P. Warners.
\newblock A linear-time transformation of linear inequalities into conjunctive
  normal form.
\newblock {\em Information Processing Letters}, 1968.

\end{thebibliography}

\end{document}